\documentclass[runningheads]{llncs}

\usepackage[utf8]{inputenc}

\usepackage{graphics} 
\usepackage{epsfig} 
\usepackage{times} 
\usepackage{amsmath} 
\usepackage{amssymb}  
\usepackage{multirow}
\usepackage{lineno}
\usepackage{tikzpagenodes}
\usepackage{multirow}
\usepackage{bigdelim}
\usepackage{subfigure}
\usepackage{float}
\usepackage{physics}
\usepackage{hyperref}

\usepackage[dvips]{rotating} 
\usepackage{array}
\usepackage{multirow}
\usepackage{subfigure}
\usepackage{rotating}

\usepackage{textcomp}
\usepackage{epsfig}
\usepackage{changes}

\title{MixUp-MIL: A Study on Linear \& Multilinear Interpolation-Based Data Augmentation for Whole Slide Image Classification}

\author{Michael Gadermayr$^1$ \and Lukas Koller$^1$ \and Maximilian Tschuchnig$^1$ \and Lea Maria Stangassinger$^2$ \and Christina Kreutzer$^3$ \and Sebastien Couillard-Despres$^3$ \and Gertie Janneke Oostingh$^2$ \and Anton Hittmair$^4$}

\authorrunning{M. Gadermayr et al.}

\institute{
$^1$ Salzburg University of Applied Sciences, Dept. of Information Technology and Digitalization\\
$^2$ Salzburg University of Applied Sciences, Dept. of Health Sciences\\
$^3$ Spinal Cord Injury and Tissue Regeneration Center Salzburg, Research Institute of Experimental Neuroregeneration\\
$^4$ Kardinal Schwarzenberg Klinikum, Department of Pathology and Microbiology
}

\begin{document}

\maketitle

\begin{abstract}
	For classifying digital whole slide images in the absence of pixel level annotations, multiple instance learning methods are typically applied. Such methods are currently of very high interest in the research community due to the generic applicability, whereby the issue of data augmentation in this context is rarely explored.
	Here, we investigate the linear and multilinear interpolation between feature vectors, a data augmentation technique which proved to be capable of improving the generalization performance of classification networks and recently also multiple instance learning.
	Recent experiments on multiple instance learning were thus far only performed on two rather small data sets in combination with one specific feature extraction approach. In this case a strong dependence on the data set could be identified. 
	We conducted a larger study incorporating $10$ different data set configurations, two different feature extraction approaches (supervised and self-supervised), stain normalization and two multiple instance learning architectures.
	The results showed an extraordinarily high variability in the effect of the method used.
\end{abstract}


\section{Introduction}\label{sec:introduction}

The trend towards whole slide imaging for digitizing histo- or cytologic slides in research and clinical practice boosts the demand for automated processing techniques~\cite{Niazi2019}. Due to the huge resolution of the whole slide images (WSIs), a holistic classification based on conventional convolutional neural network architectures is not feasible, even with state-of-the-art hardware.

Multiple instance learning (MIL)~\cite{Ilse18a,Li21a,Wang2021,Zhang2022,Shao2021,Wu2023,Wang2023,Ren2023,Fillioux2023} represents a methodology for handling these huge digital WSIs corresponding to global "image-level" labels.
In MIL, WSIs correspond to labeled bags, whereas extracted patches correspond to unlabeled instances.
Based on randomly sampling patches, MIL can be applied to WSIs showing an arbitrary size. 
MIL algorithms typically consist of a feature extraction stage, a pooling stage and a downstream classification stage.
State-of-the-art approaches mainly rely on convolutional neural networks (CNNs) for feature extraction, often in combination with attention~\cite{Li21b,Li21a} or self-attention~\cite{Rymarczyk2021}.
For training the feature extraction stage, supervised (often pre-trained) as well as self-supervised learning is employed~\cite{Li21b,Li21a}.
Most methods rely on separate individual learning stages (separation between representation learning and classification). Nevertheless, also end-to-end deep learning approaches have been proposed~\cite{Chikontwe2020,Sharma2021}.

Multiple instance learning can be categorized into instance-based and embedding-based MIL.
In the case of instance-based MIL, the information per patch is first condensed to a single scalar value, representing the classification per patch. Downstream, these patch-based values are aggregated. 
In the case of embedding-based MIL, a distinct feature vector is extracted per patch. Downstream, all feature vectors from a WSI are aggregated. Finally, the pooled embeddings are used to classify a global label.
Embedding based MIL is typically more effective in the case of a WSI-classfication, while instance-based MIL can be effectively used to intuitively detect relevant regions-of-interest (since the algorithm determines a single label per patch).
State-of-the-art models combine both, instance- and an embedding-based pathways~\cite{Li21a,Ren2023,Wang2023}. 
Zhang et al. proposed a method including the generation of multiple so-called pseudo-bags for each WSI to augment the amount of data~\cite{Zhang2022}.
Wu et al. proposed so-called smooth attention maps, incorporating the fact that images show spatial correlations~\cite{Wu2023}.
Fillioux et al. developed an efficient method for modelling large sequences (i.e. a large number of patches) when using recurrent neural networks or transformers~\cite{Fillioux2023}.
For a survey on multiple instance learning methods for classifying digital pathology, we refer to ~\cite{Gadermayr22a}.

Although the overall amount of data contained in a WSI is huge, the number of labeled samples in MIL (represented by the number of WSIs) is often small. Often, datasets are also imbalanced, further increasing this small data issue. Often, datasets are also imbalanced, further increasing this small data issue~\cite{Galdran2021}. 
General data augmentation strategies, such as rotations, flipping, stain augmentation and normalization are applicable to increase the amount of data~\cite{Tellez2019}.
All of these augmentation approaches are performed in the image domain.
Although recently there is a very high effort on improving multiple instance learning approaches~\cite{Fillioux2023,Wu2023,Wang2023,Ren2023,Liu2023,Lin2023}, there is little work on data augmentation dedicated using characteristics unique to the MIL domain.

In this work, we consider feature-level data augmentation directly applied to the patch-representations extracted using CNNs.
These methods can be easily combined with conventional image-based augmentation. The operations are computationally highly efficient and can be precomputed before starting to train the MIL downstream classification network (in the case that the tasks are separated, which often holds true).
There has been relatively little work on such forms of data augmentation.
Li et al.~\cite{Li21b} proposed an augmentation strategy based on sampling the patch-descriptors to generate several different bags for each individual WSI.
In this paper, we focus on the interpolations of patch descriptors based on the idea of Zhang et al~\cite{Zhang2017a}, which is also known as MixUp.
This method was originally proposed as data agnostic which also shows good results if applied to raw image data~\cite{Dabouei2021,Chen2022,Thulasidasan2019,Psaroudakis2020}.
Variations were proposed, to be applied to latent representations~\cite{Verma2019} and to balance data sets~\cite{Galdran2021}.

\noindent
The contribution of this paper is manifold.
\begin{itemize}
	\item We perform a large study based on our previous conference paper \cite{Gadermayr23a}, where the basic concepts of MixUp data augmentation~\cite{Zhang2017a} in WSI classification is proposed and evaluated on two small data sets.
	\item We investigate different feature extraction methods, including pretrained networks and self-supervised learning.
	\item Motivated by scores strongly depending on the data set, we investigated 10 different data set settings, derived from 3 different data sets. 	
	\item We also investigated the effect of variably restricting the data set sizes.
	\item To investigate whether stain variability plays a vital role, we analyze stain normalization in combination with data augmentation and multiple instance learning.
	\item Finally, to improve the interpretability of the results, we analyzed the variability within and between whole slide images.
\end{itemize}

\section{Methods}
In this paper, we propose and evaluate different strategies to perform data augmentation in the MIL setting, based on the pair-wise interpolation of feature descriptors.

We consider MIL approaches relying on separately trained feature extraction and classification stages~\cite{Li21a,Lerousseau21a,Rymarczyk2021}.
The proposed and evaluated data augmentation methods are applied to the patch descriptors obtained by the feature extraction stage.
This is highly efficient during training since the features are only computed once (per patch) and for augmentation only simple arithmetic operations are applied to the feature vectors.
Other generally used image-based data augmentation strategies (such as stain-augmentation, rotations or deformations) can be combined easily with the feature-based approaches but require individual feature extraction during training. To avoid the curse of meta-parameters and thereby experiments such approaches are explicitly not considered here.

Zhang et al.~\cite{Zhang2017a} originally proposed the MixUp data augmentation method, based on the idea that synthetic samples $\boldsymbol{x'}$ are generated such that
$\boldsymbol{x'}  =\alpha \cdot \boldsymbol{x_i} + (1 - \alpha) \cdot \boldsymbol{x_j} \; ,$ 
where $\boldsymbol{x_i}$ and $\boldsymbol{x_j}$ are randomly sampled raw input feature vectors. 
Corresponding labels $\boldsymbol{y'}$ are generated for a new synthetic sample such that
$\boldsymbol{y'} = \alpha \cdot \boldsymbol{y_i} + (1 - \alpha) \cdot \boldsymbol{y_j} \; ,$ 
where $\boldsymbol{y_i}$ and $\boldsymbol{y_j}$ are the corresponding one-hot label encodings. 
The weight $\alpha$ is drawn from a uniform distribution between $0$ and $1$. This can be interpreted as linear interpolation between two feature vectors.

One single input sample (corresponding to one WSI) of a MIL approach with a separate feature extraction stage~\cite{Li21a} can be expressed as a P-tupel $X = (\boldsymbol{x_1}, ..., \boldsymbol{x_P})$ with $\boldsymbol{x_i}$ being the feature vector of an individual patch and $P$ being the number of patches per WSI. 
The method proposed by Zhang et al. cannot directly be applied to this tupel representation. However, there are several options to adapt the basic idea to the changed setting.

\begin{figure}[tb] \center
	\includegraphics[width=0.99\linewidth]{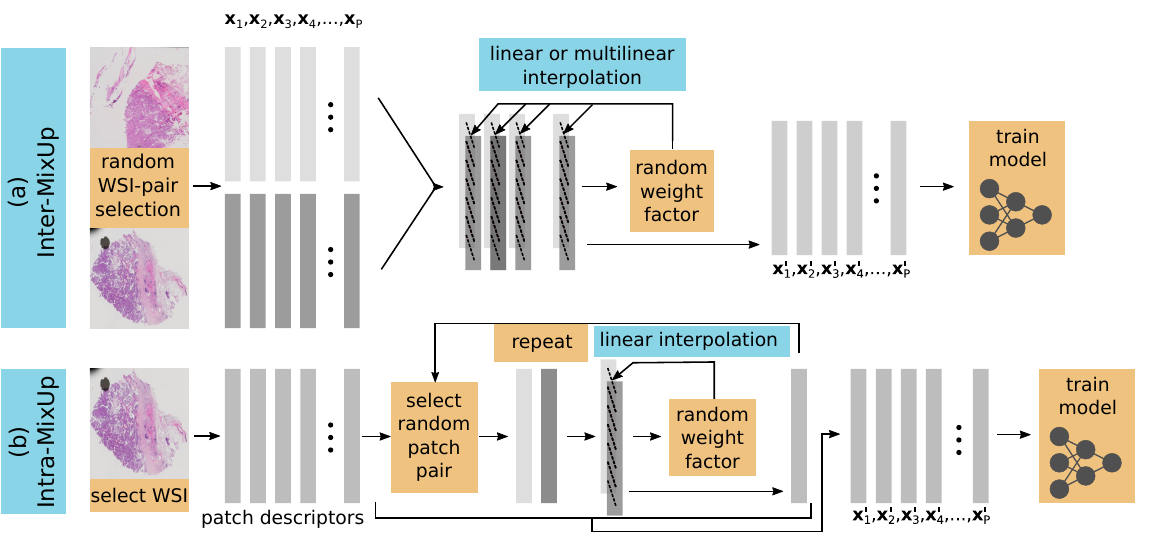}
	\caption{Overview of the proposed feature-based data augmentation approaches. In the case of Inter-MixUp (a), a linear combination is applied on the pairs of WSI descriptors with a randomly selected weight factor. In the case of Intra-MixUp (b), patch-based descriptors from the same WSI are merged with individual random weights.}
	\label{fig:outline}
\end{figure}

\subsection{Inter-MixUp} refers to the intuitive synthesis of feature vectors by linearly combining feature vectors of a pair of WSIs (see Fig.~\ref{fig:outline}~(a)).
All features of a WSI with index $w$ can be represented by $X^{(w)}$, such that
${X^{(w)}} = (\boldsymbol{x^{(w)}_{1}}, \; ... \;, \boldsymbol{x^{(w)}_{P}}) \; .$ 
\noindent
To generate a novel synthetic sample ${X^{(u)}}'$ based on two real samples ${X^{(w)}}$ and ${X^{(v)}}$, we introduce the operation 
$${X^{(u)}}' = 
(\alpha \cdot \boldsymbol{x_1^{(w)}} + (1-\alpha) \cdot \boldsymbol{x^{(v)}_1}, \; \alpha \cdot \boldsymbol{x^{(w)}_2}+ (1-\alpha) \cdot \boldsymbol{x^{(v)}_2}, \; ... \;, \;  \alpha \cdot \boldsymbol{x^{(w)}_P} + (1-\alpha) \cdot \boldsymbol{x^{(v)}_P}) $$
with $\alpha$ being a uniformly sampled random weight ($\alpha \in [0,1]$).
The WSI indexes $v$ and $w$ are uniformly sampled from the set of indexes. The index $u$ ranges from $1$ to the number of extracted WSI descriptors. This random sampling strategy theoretically allows to sample an arbitrary number of combinations in each epoch. However, since the new synthetic descriptors are individually generated in each epoch, there is no benefit of increasing the number of extracted WSI descriptors. We fix this number to the number of WSIs in the training data set, in order to keep the number of training iterations per epoch consistent over all experiments.

Two different configurations can be considered. The interpolation between WSIs of the same class (V1) and the interpolation between any WSIs, including the interpolation between different labeled WSI (V2).
In the case of V2, also the one-hot-encoded label vectors are linearly combined, such that
$\boldsymbol{y^{(u)}}' = \alpha \cdot \boldsymbol{y^{(w)}} + (1-\alpha) \cdot \boldsymbol{y^{(v)}}$ 
This leads to continuous labels.
The random values, $\alpha$, $v$ and $w$ are selected individually for each individual WSI and each epoch. 
Before applying the MixUp operation, the vector tupel containing all patch-level feature vectors, is randomly shuffled. This permutation is not specific to this data augmentation approach but is performed in each experiment.

\subsection{Intra-MixUp} 
Intra-MixUp combinations refer to the generation of synthetic descriptors by combining feature vectors within an individual WSI (see Fig.~\ref{fig:outline}~(b)).
A novel synthetic patch-level image descriptor $\boldsymbol{x_k}'$ is created based on the randomly selected descriptors $\boldsymbol{x_i}$ and $\boldsymbol{x_j}$, such that
$\boldsymbol{x_k}'= \alpha \cdot \boldsymbol{x_i} + (1-\alpha) \cdot \boldsymbol{x_j} \; , $ 
with $i$ and $j$ being random indices (uniformly sampled from $\{1, 2, ..., P\}$) and $\alpha$ being a uniformly sampled random value a ($\alpha \in [0,1]$). The index $k$ ranges from 1 to the number of extracted descriptors per patch.
The thereby obtained vector tupel $(\boldsymbol{x_1}', ..., \boldsymbol{x_P}')$ represents the synthetic WSI-level descriptor.
Besides performing combinations for each WSI during training, selective interpolation can be useful to keep real samples within the training data. This can be easily achieved by choosing 
$(\boldsymbol{x_1}', ..., \boldsymbol{x_P}')$ 
with a chance of $\beta$ and $(\boldsymbol{x_1}, ..., \boldsymbol{x_P})$ otherwise. 
While the Intra-MixUp method described before represents a linear interpolation method, with a simple adjustment also multilinear interpolation can be performed. This is achieved by computing $\boldsymbol{x_k}'$ such that 
$\boldsymbol{x_k}'= \boldsymbol{\alpha} \circ \boldsymbol{x_i} + (1-\boldsymbol{\alpha}) \circ \boldsymbol{x_j}$ with $\boldsymbol{\alpha}$ being a random vector and $\circ$ being the element-wise product. This element-wise linear (multilinear) approach enables higher variability in the generated samples.

\subsection{Experimental Architecture}

As experimental architecture, we make use of the dual-stream MIL approach proposed by Li et al~\cite{Li21a}.
Since this model combines both, embedding-based and an instance-based encoding, the effect of both paths can be individually investigated without changing any other architectural details. Since the method represents a state-of-the-art approach, it additionally serves as well-performing baseline.
In the investigated model~\cite{Li21a} an instance- and an embedding-based pathway are employed in parallel and merged by weighted addition. 
The embedding-based pathway contains an attention mechanism
, to higher weight patches that are similar to the so-called critical instance. The model makes use of an individual feature extraction stage.
\subsection{Feature Extraction}
We performed two different approaches for patch-level feature extraction.
Firstly, we extract features using a ResNet18, pre-trained on the image-net challenge data, due to the high performance in previous work and generic applicability~\cite{Gadermayr21a}.
Secondly, we trained and applied SimCLR~\cite{Chen2020a} as state-of-the-art self-supervised learning model to particularly incorporate the peculiarities of the data in spite of the absence of labels. As backbone, as in the pretrained scenario, ResNet18 was applied.
This model was assessed as particularly appropriate due to the rather low dimensional output (512 dimensions).

\subsection{Stain Normalization}
We made use of the Vahadane stain normalization technique~\cite{Vahadane2016}. 
The reference tile was chosen from the mean stain vectors of only the training data set, and both training and testing were then aligned to the reference style.

\subsection{Other Augmentation Techniques}
For a comparison with several other augmentation methods on feature level including random sampling, selective random sampling and random noise, we refer to our related study~\cite{Gadermayr23a}.
Since none of the augmentation settings based on single images lead to clear accuracy improvements, we decided to not further investigate this category of methods.

\subsection{Data Sets} \label{sec:dataset}

\subsubsection{Thyroid Paraffin \& Frozen Data Sets}
The goal of this data set~\cite{Gadermayr23a} is to distinguish different nodular lesions of the thyroid, focusing especially on benign follicular nodules (FN) and papillary carcinomas (PC).
This differentiation is crucial, due to the different treatment options, in particular with respect to the extent of surgical resection of the thyroid gland~\cite{Xi2022}.
The data set utilized in the experiments consists of 80 WSIs overall. One half (40) of the data set consists of frozen and the other half (40) of paraffin sections~\cite{Gadermayr21a}), representing the different modalities. All images were acquired during clinical routine at the 
Kardinal Schwarzenberg 
Hospital. Procedures were approved by the ethics committee of the county of Salzburg (No. 1088/2021). The mean and median age of patients at the date of dissection was 47 and 50 years, respectively. The data set comprised 13 male and 27 female patients, corresponding to a slight gender imbalance.
They were labeled by an expert pathologist with over 20 years experience. A total of 42 (21 per modality) slides were labeled as papillary carcinoma while 38 (19 per modality) were labeled as benign follicular nodule.
For the frozen sections, fresh tissue was frozen at $-15^\circ$ Celsius, slides were cut (thickness 5 $\mu m$) and stained immediately with hematoxylin and eosin.
For the paraffin sections, tissue was fixed in $4 \; \%$ phosphate-buffered formalin for 24 hours. Subsequently formalin fixed paraffin embedded tissue was cut (thickness 2 $\mu m$) and stained with hematoxylin and eosin.
The images were digitized with an Olympus VS120-LD100 slide loader system. Overviews at a 2x magnification were generated to manually define scan areas, focus points were automatically defined and adapted if needed. Scans were performed with a 20x objective (corresponding to a resolution of 344.57 nm/pixel). The image files were stored in the Oympus vsi format based on lossless compression.

Beyond the original data sets denoted as 'Frozen' and 'Paraffin', we considered the aggregation of both, denoted as 'All', since both data sets show the same biological structure. In addition, we considered stain normalization of the original data sets ('Frozen Normalized', 'Paraffin Normalized').

\subsubsection{Camelyon Data Set}
The Camelyon17 challenge data contains WSIs of hematoxylin and eosin stained lymph node sections.
We make only use of the patient-level ground truth.
All annotations were carefully prepared under supervision of expert pathologists. For the purpose of revising the slides, additional slides stained with cytokeratin immunohistochemistry were utilized. 
The data set for CAMELYON17 is collected from 5 medical centres in the Netherlands. WSI are provided as TIFF images. We differentiate between samples showing tumor and healthy samples (WSIs). We make use of 161 (normal) and 200 (patient) annotated samples\footnote{\url{http://gigadb.org/dataset/100439}}. This larger data set was further investigated to assess the effect of the data set size, with respect to both, the number of WSIs and the number of patches per WSI. Beyond the maximum number of training samples (WSIs), we reduced to 64, 32 and 16 samples per class (referred to as T64, T32, T16 in the results section). In addition, we also investigated the effect of reducing the number of patches from 2048 to 512 (P512).

\subsection{Evaluation Details} \label{sec:setup}
The data set was randomly separated into training ($80 \; \%$) and test data ($20 \; \%$). 
The whole pipeline, including the separation, was repeated 32 times to achieve representative scores. 
Due to the almost balanced setting, the overall classification accuracy (mean and standard deviation) is finally reported.
Adam was used as optimizer. The models were trained for 200 epochs with an initial learning rate of 0.0002.
Random shuffling of the vector tupels (shuffling within the WSIs) was applied for all experiments.
The patches were randomly extracted from the WSI, based on uniform sampling in informative areas. 
To obtain the informative areas, we applied thresholds based on patches extracted from downscaled slides (factor $2^{-3}$). We applied thresholding of the green color channel so ensure that large areas (at least 75 \%) are covered by tissue. In addition, we applied an entropy threshold to exclude areas showing large blurry artifacts. Due to the high image quality (without any blurry regions), the second threshold was omitted in the case of the thyroid cancer data set.
To obtain a representation independent of the WSI size, we extracted 1024 samples (thyroid data sets) and 2048 samples (Camelyon data set, due to the larger image size) patches with a size of $256 \times 256$ pixel per WSI, resulting in 1024 or 2048 patch-descriptors per WSI~\cite{Gadermayr21a}.
\href{https://gitlab.com/mgadermayr/mixupmil}{\url{https://gitlab.com/mgadermayr/mixupmil}}. 
We use the reference implementation of the dual-stream MIL approach~\cite{Li21a}.
To obtain further insight into the feature distribution, we randomly selected patch descriptor pairs and computed the cosine distances. In detail, we selected 10,000 pairs 
(a) from different WSIs with dissimilar classes only, 
(b) from different WSIs with similar and dissimilar classes, 
(c,d) from different WSIs with similar classes only, and
(e) from the same WSI.

\subsection{Source Code and Data}
The experiments are implemented using Pytorch based on a single GPU (4 GB of memory is sufficient). We make use of the reference implementation of the dual-stream MIL model~\cite{Li21a}. Source code as well as data in the form of precomputed features is available via GitLab: \url{https://gitlab.com/mgadermayr/mixupmil}.

\section{Results} \label{sec:results}

\subsection{Intra-MixUp}
\begin{figure}
\includegraphics[width=\linewidth]{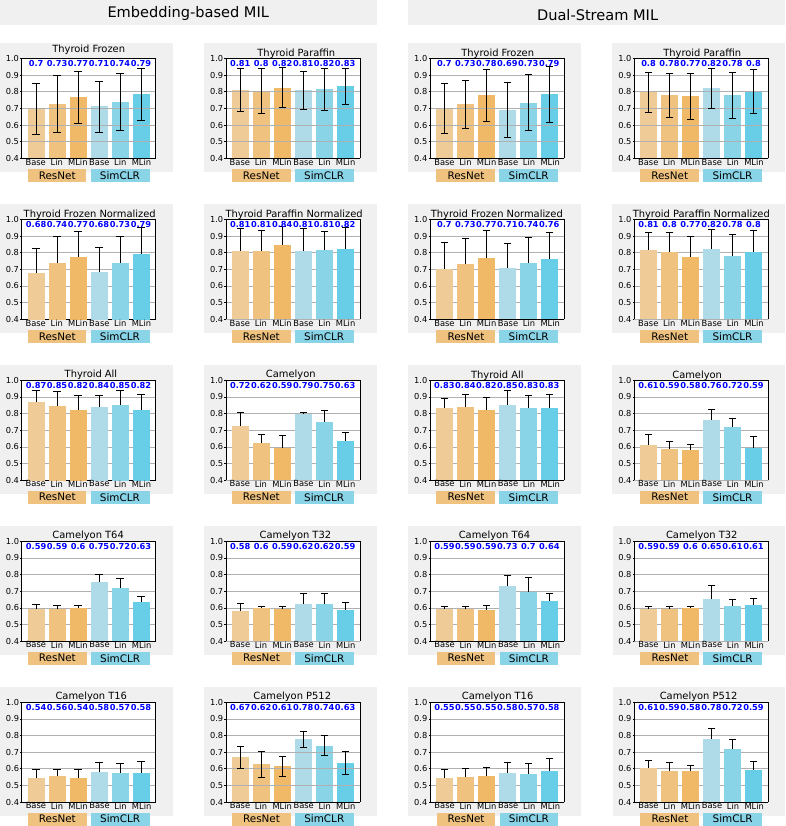}
\caption{Mean overall classification accuracy and standard deviation obtained with each individual Intra-MixUp setting. The left two columns represent embedding-based MIL, while the right column shows the scores obtained with dual-stream MIL. Each subplot shows the accuracies obtained without augmentation (Base), with linear (Lin) and multi-linear (MLin) augmentation for ResNet (left) and SimCLR (right) for one individual data set.}
\label{fig:res}
\end{figure}

Figure~\ref{fig:res} shows the mean overall classification accuracy and standard deviation obtained with each individual Intra-MixUp setting. The left two columns represent embedding-based MIL, while the right column shows the scores obtained with dual-stream MIL. Each subplot shows the accuracies obtained without augmentation (Base), with linear (Lin) and multi-linear (MLin) augmentation for ResNet (left) and SimCLR (right) for one individual data set.

Linear interpolation improved the accuracy in the case of 8 out of 16 settings with the pretrained ResNet as feature extractor.
Linear interpolation improved the accuracy in 7 out of 16 settings with the SimCLR feature extractor.
Multilinear interpolation improved the accuracy in the case of 10 out of 16 settings with the pretrained ResNet as feature extractor and in 8 out of 16 settings with the SimCLR feature extractor. In 12 out of 15 cases where linear interpolation improved accuracy, multilinear interpolation showed even larger improvements.

The largest accuracy increases with linear and multilinear interpolation where $+0.06$ (linear, Thyroid Frozen Normalized, ResNet, Embedding-based MIL) and $+0.11$ (multilinear, Thyroid Frozen Normalized, SimCLR, Embedding-based MIL).

The largest accuracy decreases with linear and multilinear interpolation where $-0.06$ (linear, Camelyon, Dual-Stream MIL, SimCLR) and $-0.17$ (multilinear, Camelyon, Dual-Stream MIL, SimCLR).

\subsection{Inter-MixUp}

\begin{figure}
	\includegraphics[width=\linewidth]{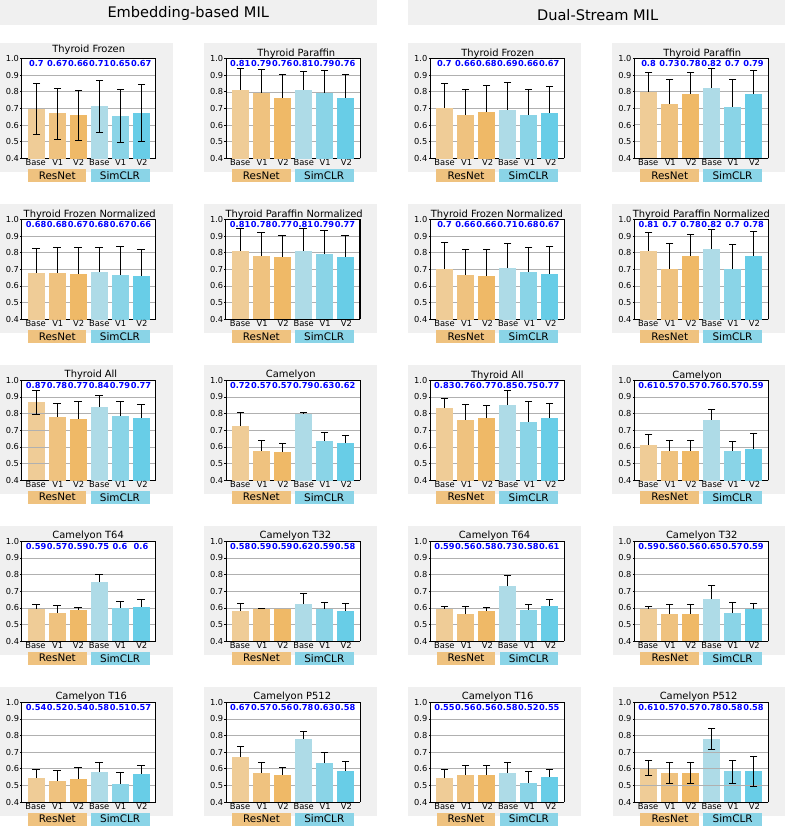}
\caption{Mean overall classification accuracy and standard deviation obtained with each individual Inter-MixUp setting. The left two columns represent embedding-based MIL, while the right column shows the scores obtained with dual-stream MIL. Each subplot shows the accuracies obtained without augmentation (Base), with linear (Lin) and multi-linear (MLin) augmentation for ResNet (left) and SimCLR (right) for one individual data set.}
\label{fig:res_inter}	
\end{figure}

Figure~\ref{fig:res_inter} shows the mean overall classification accuracy and standard deviation obtained with each individual Inter-MixUp setting. As shown in Fig.~\ref{fig:res}, the left two columns represent embedding-based MIL, while the right column shows the scores obtained with dual-stream MIL. Each subplot shows the accuracies obtained without augmentation (Base), with Inter-MixUp V1 and Inter-MixUp V2 augmentation for ResNet (left) and SimCLR (right) for one individual data set.

Marginal improvements (below $0.02$) were obtained in 5 out of 80 cases. In all of these cases, ResNet was used for feature extraction and a Camelyon-based data set was employed.

Overall, SimCLR showed better accuracies compared to the pretrained ResNet's features. In 16 out of 20 cases, the SimCLR baseline showed the higher scores. In the case of multilinear (intra) interpolation, also in 16 out of 20 cases, the best scores are obtained with SimCLR.

\subsection{Distance Analysis}
\begin{figure}[tb]
	\includegraphics[width=\linewidth]{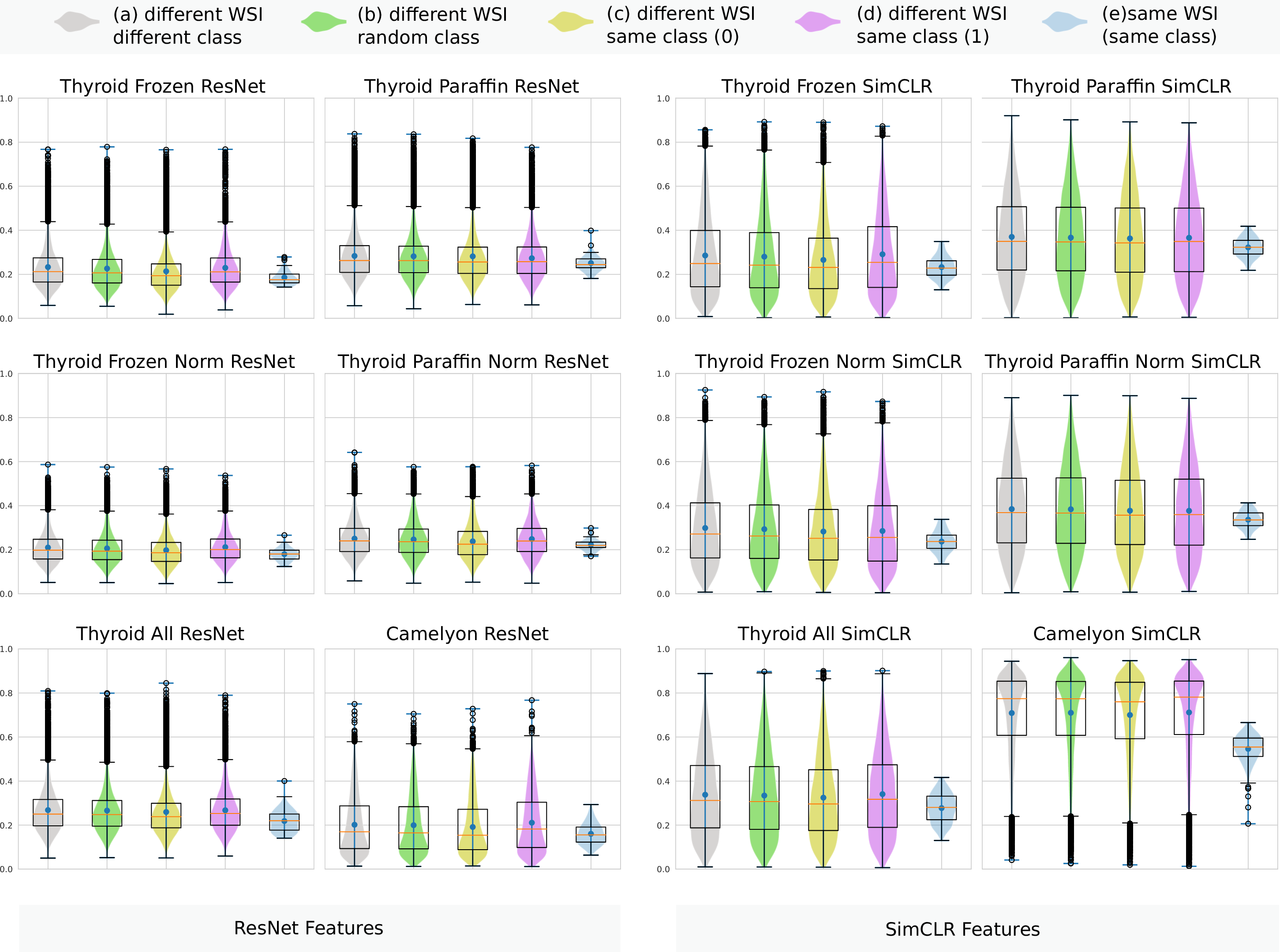}
    \caption{Analysis of the distributions of the patch descriptors' cosine distances between (a) patches from different classes, (b) randomly selected patches from different WSIs, (c,d) patches from the same class and different WSIs (for both classes, PC and FN) and (e) patches within the WSIs.}
    \label{fig:boxviolin}
\end{figure}

In Fig.~\ref{fig:boxviolin}, the distributions of the descriptor cosine distances between (a-d) patches from different WSIs (inter-WSI) and (e) patches within a single WSI (intra-WSI) are provided. 
In each setting, the intra-WSI distances (e) show the lowest values compared to similar values for the inter-WSI setting (a-d).
SimCLR shows higher values compared to ResNet features. The highest distances are obtained for the Camelyon SimCLR data set which corresponds to the data set, the self-supervised learning approach was trained on. 
Based on the used common box plot variation (whiskers length is less than $1.5 \times$ the interquartile range), a large number of data points was identified as outliers. However, these points are not considered as real outliers, but occur due to the asymmetrical data distribution (as indicated by the violin plot in the background).

\section{Discussion}

In this work, we proposed and examined novel data augmentation strategies based on the idea of interpolations of feature vectors in the MIL setting. 
Particularly, we focused on interpolating patch descriptors between (Intra-MixUp) and within whole slide images (Inter-MixUp).

The Inter-MixUp approach did not show any systematic improvements. Independently of the chosen strategy (V1, V2), the Inter-MixUp approach did not show any systematic improvements.
Since we expected that the large variability between whole slide images might negatively influence an interpolation, we performed experiments in combination with stain normalization. 
However, even with stain normalization, we did not notice any improvements.
Due to an improved feature invariance achieved with the self-supervised learning method (implicitly learning e.g. stain normalization), we also expected a positive effect of Inter-MixUp in combination with the SimCLR approach.
When comparing the scores of ResNet and SimCLR, however, we noticed that the trend is again very similar and does not strongly depend on the feature extraction technique.

The Intra-MixUp approach showed improvements in the case of several configurations. Particularly in the case of the thyroid cancer classification data sets, we noticed systematic improvements.
Independent of the feature extraction method and whether stain normalization was performed, improvements where obtained in the majority of cases. Only the combination of dual-stream MIL with the paraffin dataset showed slightly decreased scores with both, ResNet and SimCLR features. 
An explanation could be that the dual-stream approach is supposed to perform particularly well on small and challenging data sets due to the combined loss function.
Using our proposed data augmentation strategy, we were able to close the gap with the more basic embedding-based MIL approach or even outperform the state-of-the-art dual stream approach.

In the case of the Camelyon dataset, we did not notice improvements, even with clearly reduced data (see T64, T32, T16 and P512 configurations). We expected that the size of the training data would have a strong impact on the performance of data augmentation, since the availability of sufficient data was supposed to diminish positive effects of augmentation. However, whether the number of WSIs or the number of patches per WSI were reduced, we did not notice improvements.

Regarding the different feature extraction approaches, we noticed that SimCLR performed slightly better on average. Particularly in the case of the Camelyon data sets the gap was large. Considering the experiments with the proposed MixUp-MIL data augmentation, we notice similar effects in the case of both feature extraction methods. The potentially more invariant SimCLR features do not make data augmentation obsolete. 
The same holds true for stain normalization. We noticed very similar effects, independent whether stain normalization was performed or not.

Based on the distance analysis, we noticed that the intra-slide distances between pairs of patches are on average smaller for all data sets. There are particularly less pairs showing large distances (outliers). Obviously in the case of Inter-MixUp, interpolation between samples in a more diverse set of data is performed, which is supposed to have negative effects. However, based on this analysis, we are unable to explain why the Camelyon data set showed a lower performance with interpolation although similar properties are observed for both datasets.

Based on our analysis, we assume that the data set has a strong impact on the performance of linear or multilinear interpolation.
We assume that data set specific properties are important such as the amount of important (relevant) information (i.e. the number of relevant patches) in the WSIs. We assume that if only a small number of patches is relevant for the final classification (as in the case of the Camelyon data set), it is more likely that interpolation washes out the features by combining relevant with non-relevant information. For that reason we suggest to investigate more intelligent ways of selecting samples for interpolation. For example, paired samples could be selected based on neighborhood or feature similarity instead of a random selection of two feature vectors. The challenge in that case would be to find a good trade-off between a high variability in the virtual samples and not washing-out the relevant information. Finally an effective selection strategy could not only help the Intra-MixUp setting, but potentially also enable Inter-MixUp to increase diversity of data and accuracy even further.

\section{Conclusion}
We proposed and investigated novel data augmentation strategies based on the idea of interpolations of patch descriptors in the MIL setting.
Based on the experimental results, the multilinear Intra-MixUp setting proved to be highly effective in certain cases, while the Inter-MixUp method showed inferior scores compared to a state-of-the-art baseline.
We learned that there is a clear difference between combinations within and between WSIs with a noticeable effect on the final classification accuracy.
The negative effect in the case of Inter-MixUp is supposedly due to the high variability between the WSIs compared to a rather low variability within the WSIs.
This effect remains similar if the invariance of the desciptors is improved, either by performing normalization or by applying self-supervised learning.
We also did not notice any impact of the feature extraction methods used.
We found out that data set size, the number of patches per WSI, the feature extraction method and the stain variability are not the relevant factors that decide whether the proposed method performs well or not.
Based on our analysis and excluding these parameters, we strongly expect that the spatial distribution of relevant information is a crucial factor. Further we expect that a more intelligent selection of pairs for interpolation could increase performance in the case of data sets showing inferior performance if a random selection is applied.

\subsection*{Acknowledgement}
This work was partially funded by the County of Salzburg (KIAMed, AIBIA).


\bibliographystyle{splncs04}

\bibliography{bib.bib}

\begin{thebibliography}{10}
\providecommand{\url}[1]{\texttt{#1}}
\providecommand{\urlprefix}{URL }
\providecommand{\doi}[1]{https://doi.org/#1}

\bibitem{Chen2022}
Chen, J.N., Sun, S., He, J., Torr, P.H., Yuille, A., Bai, S.: Transmix: Attend
  to mix for vision transformers. In: Proceedings of the IEEE/CVF Conference on
  Computer Vision and Pattern Recognition (CVPR). pp. 12135--12144 (2022)

\bibitem{Chen2020a}
Chen, T., Kornblith, S., Swersky, K., Norouzi, M., Hinton, G.E.: Big
  self-supervised models are strong semi-supervised learners. Advances in
  neural information processing systems (NeurIPS)  \textbf{33},  22243--22255
  (2020)

\bibitem{Chikontwe2020}
Chikontwe, P., Kim, M., Nam, S.J., Go, H., Park, S.H.: Multiple instance
  learning with center embeddings for histopathology classification. In:
  Proceedings of the International Conference on Medical Image Computing and
  Computer Assisted Intervention (MICCAI). pp. 519--528 (2020)

\bibitem{Dabouei2021}
Dabouei, A., Soleymani, S., Taherkhani, F., Nasrabadi, N.M.: Supermix:
  Supervising the mixing data augmentation. In: Proceedings of the IEEE/CVF
  Conference on Computer Vision and Pattern Recognition (CVPR). pp.
  13794--13803 (2021)

\bibitem{Fillioux2023}
Fillioux, L., Boyd, J., Vakalopoulou, M., Courn{\`e}de, P.h., Christodoulidis,
  S.: Structured state space models for multiple instance learning in digital
  pathology. In: International Conference on Medical Image Computing and
  Computer-Assisted Intervention. pp. 594--604. Springer (2023)

\bibitem{Gadermayr23a}
Gadermayr, M., Koller, L., Tschuchnig, M., Stangassinger, L.M., Kreutzer, C.,
  Couillard-Despres, S., Oostingh, G.J., Hittmair, A.: Mixup-mil: Novel data
  augmentation for multiple instance learning and a study on thyroid cancer
  diagnosis. In: Proceedings of the Conference on Medical Image Computing and
  Computer Aided Interventions (MICCAI) (2023)

\bibitem{Gadermayr22a}
Gadermayr, M., Tschuchnig, M.: Multiple instance learning for digital
  pathology: A review on the state-of-the-art, limitations \& future potential.
  arXiv preprint arXiv:2206.04425  (2022)

\bibitem{Gadermayr21a}
Gadermayr, M., Tschuchnig, M., Stangassinger, L.M., Kreutzer, C.,
  Couillard-Despres, S., Oostingh, G.J., Hittmair, A.: Frozen-to-paraffin:
  Categorization of histological frozen sections by the aid of paraffin
  sections and generative adversarial networks. In: Proceedings of the MICCAI
  Workshop on Simulation and Synthesis in Medical Imaging (SASHIMI). pp.
  99--109 (2021)

\bibitem{Galdran2021}
Galdran, A., Carneiro, G., Ballester, M.A.G.: Balanced-{MixUp} for highly
  imbalanced medical image classification. In: Proceedings of the Conference on
  Medical Image Computing and Computer Assisted Intervention (MICCAI). pp.
  323--333 (2021)

\bibitem{Ilse18a}
Ilse, M., Tomczak, J., Welling, M.: Attention-based deep multiple instance
  learning. In: Proceedings of the International Conference on Machine Learning
  (ICML). pp. 2127--2136 (2018)

\bibitem{Lerousseau21a}
Lerousseau, M., Vakalopoulou, M., Classe, M., Adam, J., Battistella, E.,
  Carré, A., Estienne, T., Henry, T., Deutsch, E., Paragios, N.: Weakly
  supervised multiple instance learning histopathological tumor segmentation.
  In: Proceedings of the International Conference on Medical Image Computing
  and Computer Assisted Interventions (MICCAI) (2020)

\bibitem{Li21a}
Li, B., Li, Y., Eliceiri, K.W.: Dual-stream multiple instance learning network
  for whole slide image classification with self-supervised contrastive
  learning. In: Proceedings of the Conference on Computer Vision and Pattern
  Recognition (CVPR). pp. 14318--14328 (2021),
  \url{https://github.com/binli123/dsmil-wsi, accessed: 2022-03-14}

\bibitem{Li21b}
Li, Z., Zhao, W., Shi, F., Qi, L., Xie, X., Wei, Y., Ding, Z., Gao, Y., Wu, S.,
  Liu, J., et~al.: A novel multiple instance learning framework for covid-19
  severity assessment via data augmentation and self-supervised learning.
  Medical Image Analysis  \textbf{69},  101978 (2021)

\bibitem{Lin2023}
Lin, T., Yu, Z., Hu, H., Xu, Y., Chen, C.W.: Interventional bag multi-instance
  learning on whole-slide pathological images. In: Proceedings of the IEEE/CVF
  Conference on Computer Vision and Pattern Recognition. pp. 19830--19839
  (2023)

\bibitem{Liu2023}
Liu, K., Zhu, W., Shen, Y., Liu, S., Razavian, N., Geras, K.J.,
  Fernandez-Granda, C.: Multiple instance learning via iterative self-paced
  supervised contrastive learning. In: Proceedings of the IEEE/CVF Conference
  on Computer Vision and Pattern Recognition. pp. 3355--3365 (2023)

\bibitem{Niazi2019}
Niazi, M.K.K., Parwani, A.V., Gurcan, M.N.: Digital pathology and artificial
  intelligence. The lancet oncology  \textbf{20}(5),  e253--e261 (2019)

\bibitem{Psaroudakis2020}
Psaroudakis, A., Kollias, D.: Mixaugment \& mixup: Augmentation methods for
  facial expression recognition. In: Proceedings of the Conference on Computer
  Vision and Pattern Recognition Workshops (CVPRW) (2020)

\bibitem{Ren2023}
Ren, Q., Zhao, Y., He, B., Wu, B., Mai, S., Xu, F., Huang, Y., He, Y., Huang,
  J., Yao, J.: Iib-mil: Integrated instance-level and bag-level multiple
  instances learning with label disambiguation for pathological image analysis.
  In: International Conference on Medical Image Computing and Computer-Assisted
  Intervention (MICCAI). pp. 560--569. Springer (2023)

\bibitem{Rymarczyk2021}
Rymarczyk, D., Borowa, A., Tabor, J., Zielinski, B.: Kernel self-attention for
  weakly-supervised image classification using deep multiple instance learning.
  In: Proceedings of the IEEE/CVF Winter Conference on Applications of Computer
  Vision (WACV). pp. 1721--1730 (2021)

\bibitem{Shao2021}
Shao, Z., Bian, H., Chen, Y., Wang, Y., Zhang, J., Ji, X., zhang, y.: Transmil:
  Transformer based correlated multiple instance learning for whole slide image
  classification. In: Advances in Neural Information Processing Systems
  (NeurIPS). vol.~34, pp. 2136--2147 (2021)

\bibitem{Sharma2021}
Sharma, Y., Shrivastava, A., Ehsan, L., Moskaluk, C.A., Syed, S., Brown, D.:
  Cluster-to-conquer: A framework for end-to-end multi-instance learning for
  whole slide image classification. In: Proceedings of the Medical Imaging with
  Deep Learning Conference (MIDL). pp. 682--698 (2021)

\bibitem{Tellez2019}
Tellez, D., Litjens, G., Bándi, P., Bulten, W., Bokhorst, J.M., Ciompi, F.,
  {van der Laak}, J.: Quantifying the effects of data augmentation and stain
  color normalization in convolutional neural networks for computational
  pathology. Medical Image Analysis  \textbf{58},  101544 (2019)

\bibitem{Thulasidasan2019}
Thulasidasan, S., Chennupati, G., Bilmes, J.A., Bhattacharya, T., Michalak, S.:
  On mixup training: Improved calibration and predictive uncertainty for deep
  neural networks. In: Advances in Neural Information Processing Systems
  (NeurIPS). vol.~32 (2019)

\bibitem{Vahadane2016}
Vahadane, A., Peng, T., Sethi, A., Albarqouni, S., Wang, L., Baust, M.,
  Steiger, K., Schlitter, A.M., Esposito, I., Navab, N.: Structure-preserving
  color normalization and sparse stain separation for histological images. IEEE
  transactions on medical imaging  \textbf{35}(8),  1962--1971 (2016)

\bibitem{Verma2019}
Verma, V., Lamb, A., Beckham, C., Najafi, A., Mitliagkas, I., Lopez-Paz, D.,
  Bengio, Y.: Manifold mixup: Better representations by interpolating hidden
  states. In: Proceedings of the International Conference on Machine Learning
  (ICML). vol.~97, pp. 6438--6447 (2019)

\bibitem{Wang2023}
Wang, H., Luo, L., Wang, F., Tong, R., Chen, Y.W., Hu, H., Lin, L., Chen, H.:
  Iteratively coupled multiple instance learning from instance to bag
  classifier for whole slide image classification. In: International Conference
  on Medical Image Computing and Computer-Assisted Intervention (MICCAI) (2023)

\bibitem{Wang2021}
Wang, X., Yang, S., Zhang, J., Wang, M., Zhang, J., Huang, J., Yang, W., Han,
  X.: {TransPath}: Transformer-based self-supervised learning for
  histopathological image classification. In: Proceedings of the Conference on
  Medical Image Computing and Computer Assisted Intervention (MICCAI). pp.
  186--195 (2021)

\bibitem{Wu2023}
Wu, Y., Castro-Mac{\'\i}as, F.M., Morales-{\'A}lvarez, P., Molina, R.,
  Katsaggelos, A.K.: Smooth attention for deep multiple instance learning:
  Application to ct intracranial hemorrhage detection. In: International
  Conference on Medical Image Computing and Computer-Assisted Intervention. pp.
  327--337. Springer (2023)

\bibitem{Xi2022}
Xi, N.M., Wang, L., Yang, C.: Improving the diagnosis of thyroid cancer by
  machine learning and clinical data. Scientific Reports  \textbf{12}(1) (2022)

\bibitem{Zhang2022}
Zhang, H., Meng, Y., Zhao, Y., Qiao, Y., Yang, X., Coupland, S.E., Zheng, Y.:
  Dtfd-mil: Double-tier feature distillation multiple instance learning for
  histopathology whole slide image classification. In: Proceedings of the
  IEEE/CVF Conference on Computer Vision and Pattern Recognition (CVPR). pp.
  18802--18812 (2022)

\bibitem{Zhang2017a}
Zhang, H., Cisse, M., Dauphin, Y.N., Lopez-Paz, D.: mixup: Beyond empirical
  risk minimization. In: Proceedings of the International Conference on
  Learning Representations (ICLR) (2018)

\end{thebibliography}

\end{document}